\documentclass{ieeeaccess}
\pdfoutput=1
\usepackage{cite}
\usepackage{amsmath,amssymb,amsfonts}
\usepackage{algorithmic}
\usepackage{graphicx}
\usepackage{textcomp}
\usepackage{array}
\usepackage{siunitx}
\usepackage{hyperref}
\usepackage[table]{xcolor}
\def\BibTeX{{\rm B\kern-.05em{\sc i\kern-.025em b}\kern-.08em
    T\kern-.1667em\lower.7ex\hbox{E}\kern-.125emX}}
\begin{document}
% tmp comment out for arxiv
% \history{Date of publication xxxx 00, 0000, date of current version xxxx 00, 0000.}
\doi{000000000000}

\title{ISETAuto: Detecting vehicles with depth and radiance information}
\author{\uppercase{Zhenyi Liu}\authorrefmark{1},
\uppercase{Joyce Farrell\authorrefmark{2} and Brian Wandell}\authorrefmark{2}}

\address[1]{State Key Laboratory of Automotive Simulation and Control, Jilin University (e-mail: zhenyiliu27@gmail.com)} 
\address[2]{Stanford University (e-mail: {jefarrel, wandell}@stanford.edu)}
% \tfootnote{Supported by Jilin University. }

% tmp comment out for arxiv
% \markboth
% {Z.Liu \headeretal: ISETAuto: Detecting vehicles with depth and radiance information}
% {Z.Liu \headeretal: ISETAuto: Detecting vehicles with depth and radiance information}

% \corresp{Corresponding author: Zhenyi Liu (e-mail: zhenyiliu27@gmail.com)}

\begin{abstract}
Autonomous driving applications use two types of sensor systems to identify vehicles - depth sensing LiDAR and radiance sensing cameras. We compare the performance (average precision) of a ResNet for vehicle detection in complex, daytime, driving scenes when the input is a depth map (D = d(x,y)), a radiance image (L = r(x,y)), or both [D,L]. (1) When the spatial sampling resolution of the depth map and radiance image are equal to typical camera resolutions, a ResNet detects vehicles at higher average precision from depth than radiance. (2) As the spatial sampling of the depth map declines to the range of current LiDAR devices, the ResNet average precision is higher for radiance than depth. (3) For a hybrid system that combines a depth map and radiance image, the average precision is higher than using depth or radiance alone. We established these observations in simulation and then confirmed them using real-world data. The advantage of combining depth and radiance can be explained by noting that the two type of information have complementary weaknesses.  The radiance data are limited by dynamic range and motion blur.  The LiDAR data have relatively low spatial resolution. The ResNet combines the two data sources effectively to improve overall vehicle detection. 
\end{abstract}


\titlepgskip=-15pt

\maketitle

\section{Introduction}
\label{sec:introduction}
\PARstart{P}{eople} can detect vehicles using only a single eye (monocular), and people with monocular vision are permitted to drive. Moreover, people can accurately recognize objects from 2D images that contain no binocular (stereo) information. These simple observations raise a practical question: Given the high accuracy of vehicle detection using monocular information or 2D images, how much will stereo information improve accuracy? 

The reverse formulation of this question is also interesting. Bats, the second largest order of mammals after rodents, include many species that navigate through complex environments using depth sensing \cite{Griffin1965-bs,Griffin1958-co}. There are several clear advantages to using depth sensing. Under low illumination conditions, such at night and in caves, radiance data are unreliable. Further, depth sensing avoids some of the challenging aspects of radiance measurements, such as non-uniform illumination and shadows. How well can a system perform using only depth information?

This paper assesses the value of explicit depth information such as one might obtain from LiDAR systems in an automotive application. This assessment has practical significance because obtaining and using accurate depth information can be expensive. At present, LiDAR sensors for automotive imaging are a significant part of the cost of an automotive object identification system and has increases the complexity of the system. Before committing to explicit depth measurements from LiDAR, it seems sensible to quantify the benefits.

We refer to the depth information from LiDAR as explicit depth information to distinguish it from the wealth of implicit depth information in monocular images. The implicit depth information is present in the form of object size, occlusion, and texture gradients, and this information is routinely used in human visual perception \cite{E_Bruce_Goldstein_and_James_R_Brockmole2017-xi}.  The implicit depth information may be used by deep networks that identify objects, such as vehicles, from images. 

The simulations reveal great value in explicit depth information for detecting vehicles. This does not suggest that radiance data are not important: some critical driving information (road markings, traffic light status) is available through radiance sensors but not depth sensors. Therefore, we investigated a simple network architecture that combines radiance and depth information, and we explore the network’s performance with the fused sensor data.

\section{Related work}
\label{sec:related work}
A number of authors have explored the value of depth information on its own or in combination with radiance data. These groups have used neural networks with a variety of architectures. Some authors input explicit depth and imaging data on independent channels that are processed separately through several network layers \cite{Sun2020-bf,Chen2018-dj,Hazirbas2016-is,Ophoff2019-wo}.  After some processing, the depth and radiance channels are fused. This design makes it possible to use either imaging or non-imaging representations of the depth data, such as point clouds.

This design raises the question of which level is best for combining the independent channels; one might expect that the answer depends on both the network and the data. Using a YOLOv2 network, \cite{Ophoff2019-wo} explored how variations in the merge-layer influenced performance. The portions of their analysis most relevant to our work detected vehicles using data obtained from the widely-used KITTI database. The average precision performance for moderately difficult vehicles (77\%) was better than depth alone (64\%). For their case, the optimal combination was (80\%). 

Caltagirone et al. used a more complex network architecture involving the cross-fusion of two different input channels for identifying the road surface \cite{Caltagirone2019-zw}. They converted explicit depth information from point clouds into depth maps and compared various fusion strategies.  Performance on depth maps was slightly higher than performance on the RGB image data, though in this task performance was nearly at the ceiling in most cases.

Additional papers examining architectures that integrate depth and RGB data streams have been explored by multiple authors \cite{Gupta2014-ux,Hoffman2016-tt,Yuanzhouhan_Cao2017-qr}. These authors report that explicit depth information significantly improves object detection performance.

An alternative conceptualization is provided by \cite{Wang2018-vu}.  These authors focus on the representation of the depth information, showing that a transformation of the input from depth map images into 3D point cloud representations substantially improves 3D vehicle detection, which is more challenging than 2D image location (IOU).  They report obtaining performance vehicle detection on the KITTI dataset improves from 22\% to 74\%.  Note that converting from depth map to point cloud is the opposite of the representation used by Ophoff et al. (2019)\cite{Ophoff2019-wo}.

\section{Contributions}

In summary, our contributions are:

a) The existing literature relies heavily on the KITTI data set which has unique properties that make it dissimilar to other empirical data sets \cite{Liu2020-de}. We extend the literature by using image system simulation which has better generalization properties and also enables us to sweep out a much larger range of image system designs \cite{Liu2019-tc}. We implement an open-source, freely available image systems simulation to model camera images and LiDAR images in relatively complex 3D automotive scenes.

b) Second, we use a ResNet \cite{He2016-au} whose performance is generally higher than other networks \cite{Reith2019-wy}. We evaluate the ability to perform vehicle detection based on training and testing with depth maps. We further assess how performance depends on the spatial sampling of both depth and radiance information.

c) Third, we evaluate using explicit depth information either in isolation or combined with image radiance data as input to a convolutional neural network. The simplicity and widespread use of the ResNet make it an attractive and practical approach for integrating RGB and depth data at the input layer of a convolutional neural network.

d) Fourth, We validate simulations on a public Waymo data set, finding a close agreement between the simulations and real-world data analyses.

\section{Methods}
\subsection{ISETAuto simulated driving scenes}
We generated a collection of complex scenes for vehicle detection using open source software, ISET3d \cite{Liu2019-tc}. This software allows users to create scene spectral radiance from pre-built or user-assembled three-dimensional scenes rendered with physically based rendering techniques\cite{Pharr2016-po}. Lights and surface reflectances are represented as spectral quantities, and the size, position, and movement of assets are defined in physical units based on a driving simulator\cite{Behrisch2011-zw}.

We created scenes by randomly sampling assets (e.g. vehicles, buses, trucks, pedestrians, buildings, roads, trees, etc.) from a collection we maintain on a cloud-based database, Flywheel. The asset positions were defined by models of street scenes and the random movements in the driving simulator. Each scene is unique and the collection is designed to maximize the diversity of the images of daytime driving scenes. In previous work, we quantified how well training on the ISETAuto dataset generalizes to real-world datasets, including KITTI, CityScape, Baidu-Apollo, and Berkeley Deep Drive \cite{Liu2020-de}. In this paper we add new comparisons with a Waymo dataset.

\subsubsection{Simulated Camera (Radiance) dataset}
We use open-source software, ISETCam\cite{Farrell2012-wn}, to convert the scene spectral radiance to sensor voltages. We simulate an automotive sensor - MT9V024 sensor manufactured by ON Semiconductor- that provides different color filter array options for automotive vision applications, e.g. monochrome, RGB Bayer, and RCCC color filter arrays. The MT9V024 sensor has \SI{2.5}{\micro\meter} pixels with relatively high light sensitivity, high signal-to-noise and a dynamic range of 55 dB. 

\subsubsection{Simulated LiDAR (Depth) dataset}
We generated scene depth maps by tracing rays from each camera pixel into the 3D world. Rays are traced from each pixel through the principal point of the lens. When the ray reaches the surface of an asset, or the environment map, we record the [x, y, z] value. The depth information is stored for every pixel to form the depth map.

\subsubsection{Real-world dataset (Waymo)}
The Waymo dataset\cite{Sun2019-bl} consists of 1150 video sequences (20 sec) of different scenes. It comprises well registered LiDAR and camera data. Waymo provides 5 different camera views (Front, front right, front left, side right, side left). In this paper, we collected images from the front camera as the dataset for network training and evaluation. The original spatial resolution of the images is 1920x1280, however, due to the limited vertical field of view of provided LiDAR data, to match the camera and LiDAR data, we cropped the camera images to 1920x743. 

The Waymo LiDAR data have a maximum distance of 76 meters.  Thus, the LiDAR detection range is smaller than the distance encoded by the RGB images. In the following experiments, we reported the training results only based on the labeled objects up to 76 meters. We picked the images from every 100 images in the Waymo video sequences to create a diverse dataset consisting of 3700 images, 3000 images for training, 350 held-out images for testing, and another 350 held-out images for evaluation.

\subsection{Object detection network \& metrics}
We use Mask R-CNN\cite{Gkioxari2019-mw} with a ResNet50 as the backbone for vehicle detection. The performance of a ResNet network is higher than other networks we have tested \cite{Reith2019-wy}. The simplicity and widespread use of the ResNet model make it an attractive and practical system for integrating RGB and depth data. Mask R-CNN includes a region proposal network (RPN) that specifies different regions in an image where an object might be found; The fully connected layers is used for bounding box classification and regression. 

A detection is considered correct when the area of the intersection of the labeled vehicle bounding box with the proposed region is greater than 50\% of the area of the union of the proposed region and the bounding of the vehicle (intersection over union, IoU).  Combining the hits and false alarms from this measure, we obtain the average precision of the IoU, a metric that is widely used in machine-learning\cite{Everingham2010-sk}. Unless indicated otherwise, we use the shorthand AP to describe AP@0.5IoU.

We trained all models from scratch. Training was based on 3000 images which were presented to the network with a batch size of 8 images per training step; model weights were updated after each batch. For example, for the case of 40,000 training steps, a total of 320,000 images were presented and the training set of 3000 images was presented about 106 times (epochs). The model was evaluated and the AP values saved at 16 checkpoints. Model performance was evaluated based on 700 images that were not used in training (held out). We trained and evaluated the model for vehicles and pedestrians using 4 Nvidia P100 GPUs. We performed the same training and evaluation strategy for all experiments in this paper.

\section{Results}
\subsection{Equated for spatial resolution, depth is better than radiance for vehicle detection in complex scenes.}
How much information about the presence or absence of a vehicle is contained in high-quality depth measurements? We addressed this question by simulating pixel-wise depth maps from 3000 different driving scenes. We converted these depth maps into linear gray-scale image values (i.e., a depth map). We trained a ResNet to detect vehicles from a subset of these images and evaluated performance after training using the held-out data (350 images).  An example of a simulated RGB image, monochrome image, and depth map are shown in Figure \ref{fig1}. The simulated depth maps have high spatial resolution, beyond what is typically measured by LiDAR and also beyond the accuracy of explicit depth information that can be obtained from even the best stereo algorithms. 

We trained the ResNet using each of these types of simulated inputs and compared the AP@0.5IoU on held-out data after training.   The AP based on depth alone is quite high, a little more than 90\%, and substantially higher than the AP from either the RGB or monochrome cameras at matched resolution (Figure \ref{fig1}, bar plot). This simulation and many others in this paper show that a high resolution and noise-free depth map contains a great deal of information that can be used to identify the positions of the vehicles.

\Figure[t!](topskip=0pt, botskip=0pt, midskip=0pt)[width=3 in]{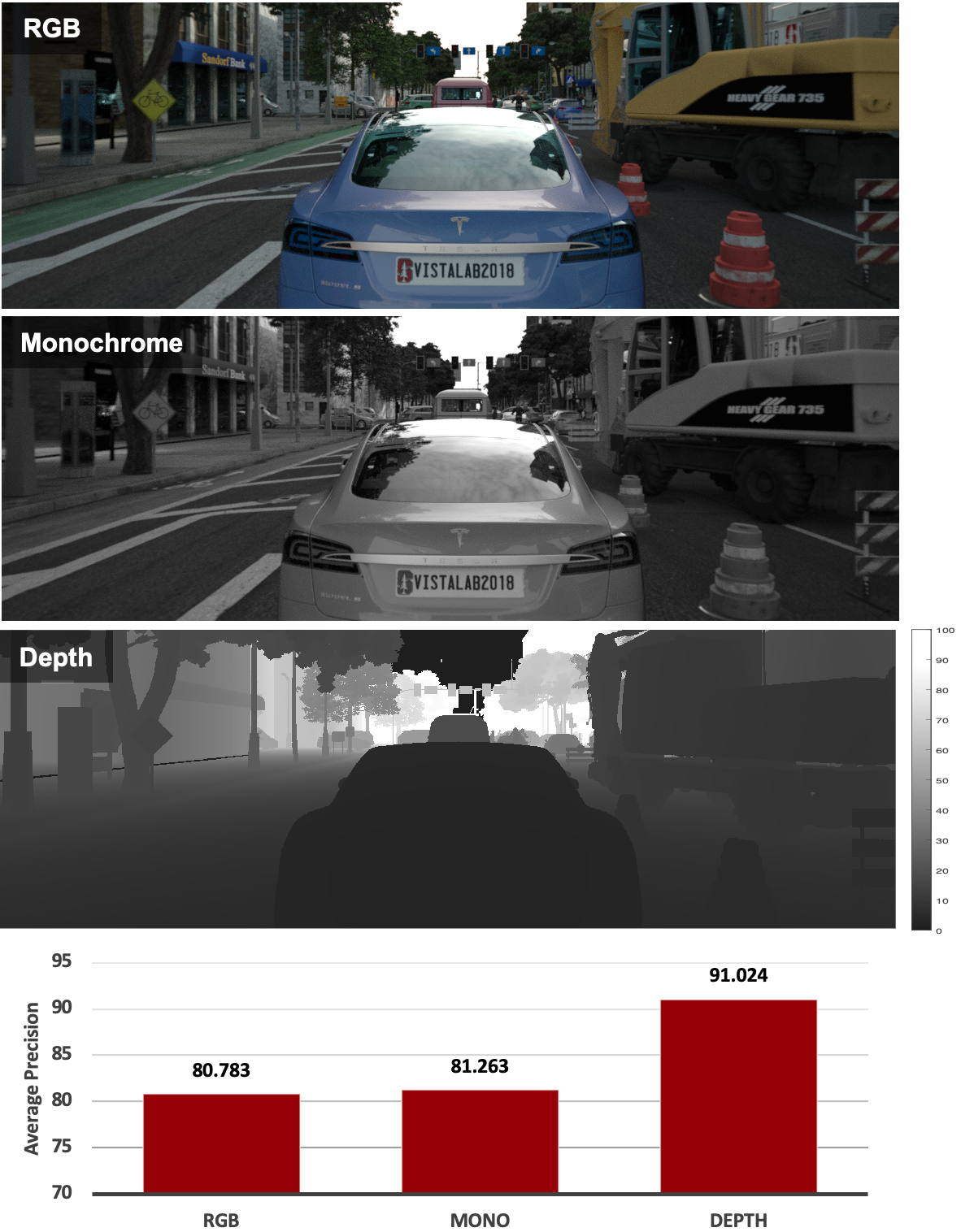}
{\ A simulated scene captured with different types of sensors: RGB, monochrome, and depth. The radiance sensor parameters are modeled on the MT9V024 sensor by ON Semiconductor, which has a \SI{2.5}{\micro\meter} pixel, 1920 x 650 resolution for a 64 x 21 deg field of view. The depth data are calculated from the simulated scene - LiDAR maps do not produce such high resolution maps. The bar chart shows the average precision (AP) for vehicle detection on networks trained using each data type. High spatial resolution depth information outperforms the RGB or monochrome sensor data by about 10\%.
\label{fig1}}

\subsection{At LiDAR spatial resolution, depth is less effective than radiance}

The scene spatial sampling from a LiDAR sensor is typically lower than the spatial sampling by a camera.  In addition, the LiDAR samples are typically matched to the angular resolution that is natural for a beam that sweeps across the scene, which corresponds to the pixel spatial sampling of a pinhole camera. The spatial sampling in a camera depends on the geometric transformation imposed by the camera optics. The examples here used a lens model with relatively little geometric distortion.

Like all physical devices, the LiDAR depth estimates include noise. The amount of noise depends on factors including the distance to the object, the incident angle of rays and the reflectance of the target. In this section, we consider the consequences of obtaining LiDAR data that are at different spatial sampling resolutions and with measurement error in the depth estimate.

\subsubsection{The dependence on spatial sampling}
We calculate the average precision for vehicle detection using spatial sampling patterns that are typical of commercial LiDAR systems \cite{Manufacturer_undated-wn}.

The panels in Figure \ref{fig2}A-C contain images of a simulated camera along with superimposed spatial sampling patterns for LiDAR systems with different horizontal and vertical resolutions. The image was modeled as if taken by a wide-angle lens; the superimposed LiDAR sample points are those we expect to obtain assuming the LiDAR images correspond to the sampling through pinhole optics model of the same scene. The lens introduces a small geometric distortion, so there will be a small (1-2 pixel) misalignment between the spatial samples of the LiDAR device and the pixel responses in the camera image.  

The chart in Figure \ref{fig3} compares the AP when networks are trained and then evaluated using only LiDAR data with different spatial sampling resolutions.  The densities range from very high, equal to the camera image (AP $\approx$ 90\%), to a more realistic density with only 0.2\% as many samples (AP $\approx$64\%). The chart shows that when the LiDAR sampling rate is reduced to only 1.5\% of the image sampling density (hor - 0.2 deg/sample and vertical = 0.33 deg/sample), performance remains quite high (AP $\approx$87\%).

\Figure[t!](topskip=0pt, botskip=0pt, midskip=0pt)[width=3.35 in]{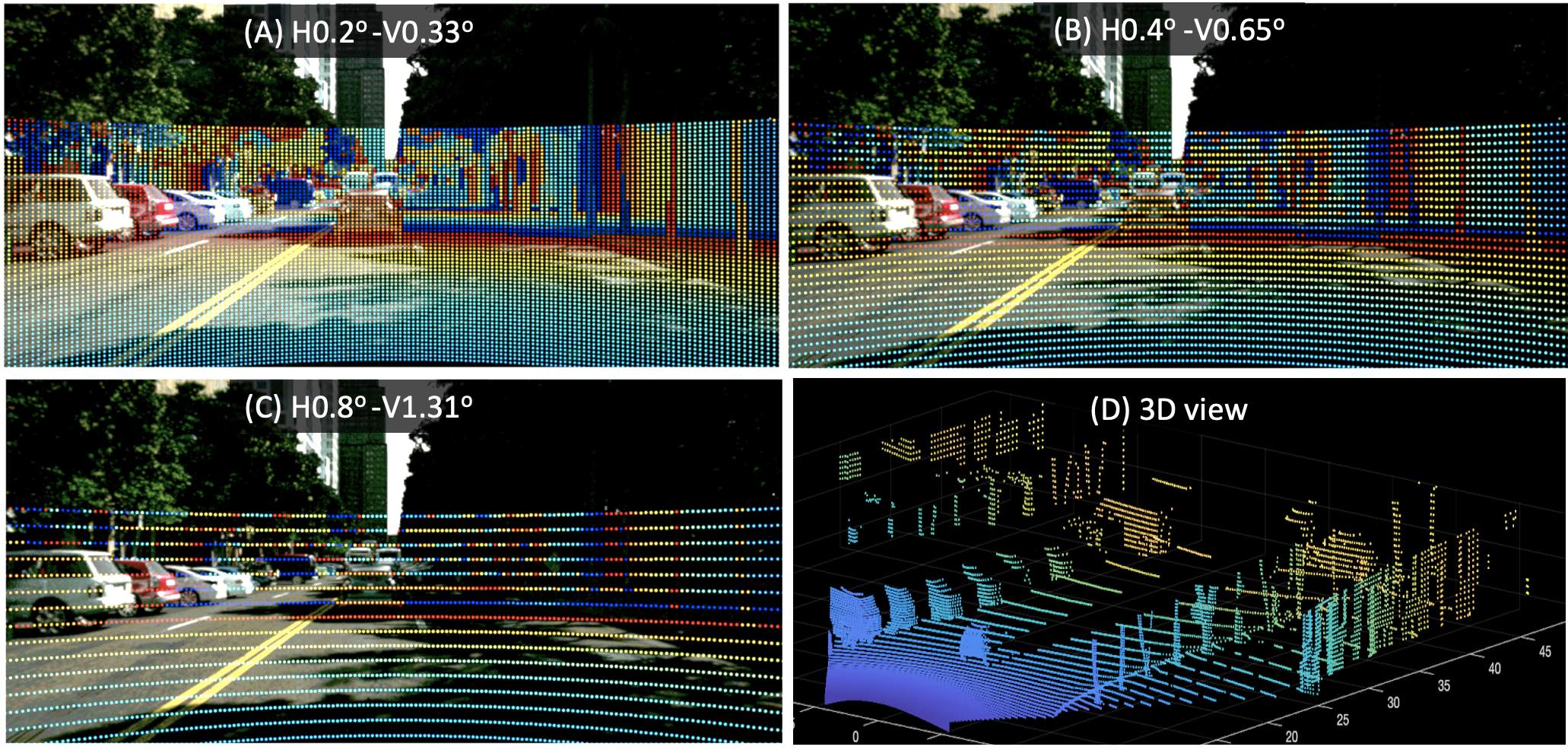}
{\ Illustration of the different LiDAR spatial resolutions used in the simulations. Panels A-C show the sampling resolution for different horizontal and vertical resolutions (deg/sample). Notice that the LiDAR sampling does not extend into upper angles because, at present, there are no flying vehicles. Panel D is a point cloud representation of the view in Panel (A). The point cloud is useful for human visualization; it contains the same information as the depth map.
\label{fig2}}

\Figure[t!](topskip=0pt, botskip=0pt, midskip=0pt)[width=3 in]{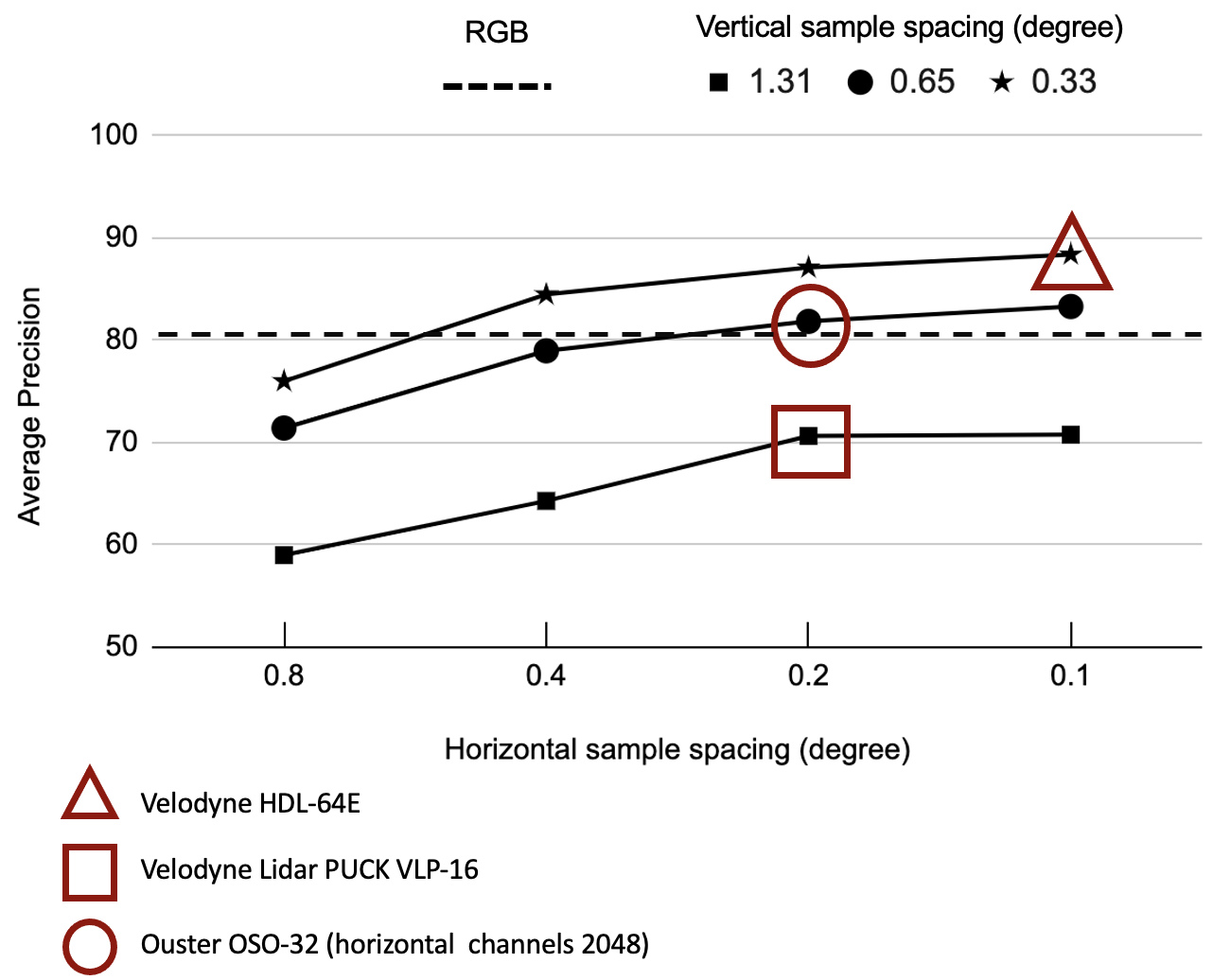}
{\ The average precision for detecting vehicles from depth data at different spatial resolutions. The lines connect simulations at a common vertical resolution, and the x-axis measures horizontal resolution.  The red symbols indicate the spatial resolution of different commercial sensors.  The performance using data from an RGB camera (1920 x 650) is shown by the horizontal dashed line.
\label{fig3}}

\subsubsection{Depth measurement error}
Next, we considered the impact of depth-measurement noise on system performance. First, we discuss the noise model, and then we report on the performance when the simulations include noise in the depth data.

LiDAR systems typically send a sequence of rays with a particular level of energy. As the ray travels from the light source to the object and returns, multiple factors that reduce the returned energy level. The factors include transmission loss in the medium, surface reflectivity, geometry relating the incident ray angle, the surface normal, and the surface bidirectional reflectance distribution function. The main source of noise in the LiDAR signal can be attributed to energy reduction. In many cases, the returned ray is not detected by the LiDAR sensor, which is dropout noise.

This noise can be modeled in three different ways. First, the noise can be simulated by randomly deleting points from the simulated sensor \cite{Fang2020-gr}. Second, in principle the noise can be simulated by quantitatively accounting for each of the factors that impact the ray transmission, keeping only the rays whose returned energy exceeds a threshold (physical simulation). This approach is somewhat impractical because so many factors are currently unspecified. Third, some investigators have trained a neural network to delete sensor data based on a set of examples of real data \cite{Manivasagam2020-lx}. 

We use the first approach: we delete a random selection of 20\% of the data points from the depth map. We use this approach because we do not have enough baseline data to characterize the different sources of physical signal loss (physical simulation), and we do not have a large amount of training data that can be used to train a neural network for the conditions we are simulating. Note that randomly deleting samples reduces the spatial resolution. Hence,it is possible to uniformly, rather than randomly, downsample the data and create an data set with the same number of samples.

\Figure[t!](topskip=0pt, botskip=0pt, midskip=0pt)[width=3 in]{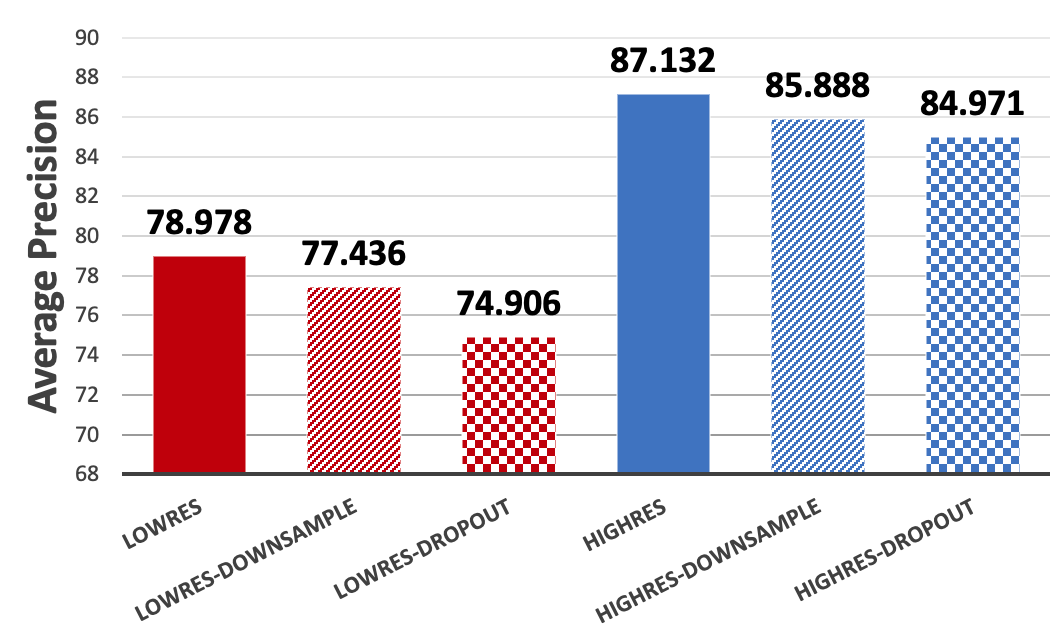}
{\ AP for vehicle detection when using different depth spatial sampling strategies. Simulations were carried at using low (red) spatial sampling (0.4 deg horizontal, 0.65 deg vertical) and  high (blue) spatial sampling (0.2 deg horizontal, 0.33 deg vertical). The sampling density was reduced by 20\% either by uniformly downsampling (cross-hatched) or by randomly deleting samples (checkerboard).
\label{fig4}}

We trained and evaluated the ResNet on noise-free data at high and low spatial sampling resolution (Figure \ref{fig4}, solid bars), and with two types of reduced spatial resolution: downsampling and dropout noise sampling (20\%). The reduction in spatial resolution was selected to match the reduction samples in the dropout noise. Uniform downsampling is slightly preferred to dropout noise for both spatial resolutions (Figure \ref{fig4}, cross-hatched and checkers).  The spatial downsampling decreases the AP by 2-3 percent at both low and high resolution. The decrease in AP the dropout noise is 3-4 percent.  

\subsection{Sensor Fusion: Combining radiance and depth maps}
Radiance and depth data have complementary strengths.  Depth information is particularly helpful under very low illumination or when the image is not properly exposed; this can easily happen in a high dynamic range scene when some objects are in direct light and the rest are in shadow. An extreme case is driving through a tunnel. In such conditions, the radiance data captured by the camera may not effectively represent both the bright and dark regions. Depth information is largely invariant to such differences in ambient illumination, making depth helpful for filling in information missed by a poorly exposed camera image.  

Furthermore, radiance data are often obtained using exposure durations that are far longer than the nearly instantaneous temporal point sampling of the LiDAR detector. When measuring nearby moving targets, the radiance data can include a significant amount of motion blur (Figure \ref{fig8}).  An advantage of radiance data is that they are easily obtained at much higher spatial resolution than LiDAR, and the data can be acquired at higher frame rates. In general, the radiance data is effective for detecting distant objects that have a small angular extent.  Radiance data is also essential for tasks such as finding road markings or identifying traffic light status.

\Figure[t!](topskip=0pt, botskip=0pt, midskip=0pt)[width=3.35 in]{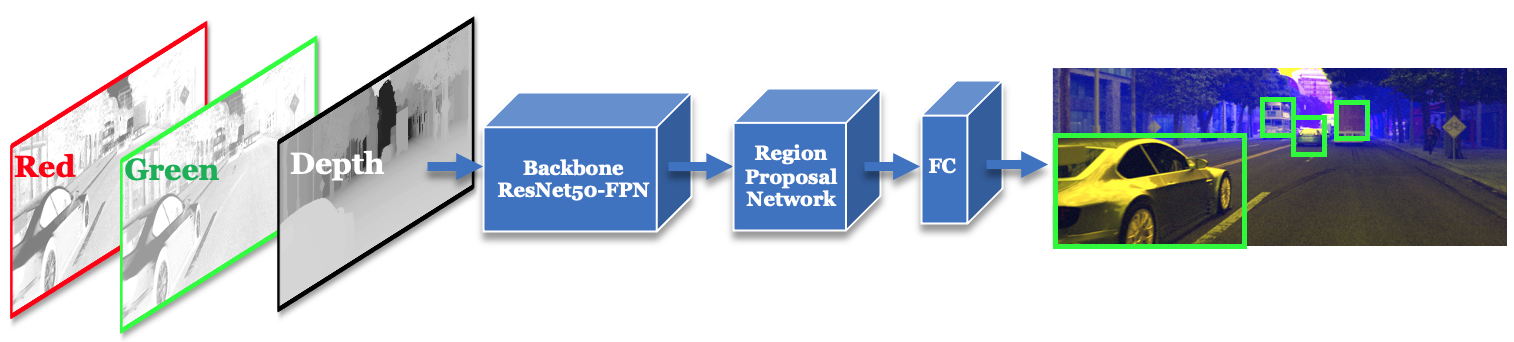}
{\ Visualization of the ResNet network architecture.  The red, green and depth images are input to a ResNet backbone, followed by a region proposal network, and a fully-connected output layer. The final image shows  regions where the network identifies vehicles (green rectangles).  The rectangles are superimposed on the merged RGD image.
\label{fig5}}

The complementary strengths of the two types of information suggest that a system that combines radiance and depth data may be particularly effective. In the section "Related Work", we describe a number of papers that explore systems that integrate radiance and depth data.  The analyses we have performed suggest that these two types of information can be combined by entering the radiance and depth map in different channels of a standard CNN input.

\subsubsection{Creating RGD inputs}
We created images by combining the R and G channels from a simulated camera with the depth channel from a simulated LiDAR device, the depth channel is normalized (0-1) and converted to 8bit integers (0-255). These RGD images have two channels at relatively high spatial resolution (RG) and one channel at lower spatial resolution (D).  In separate experiments, we filled in the missing values in the D channel with zeroes, and in other cases, we used linear interpolation to fill in the missing data.  There were no significant differences between these methods.

The RGD data are rendered as color images in Figure \ref{fig6}. The top image shows a simulated driving scene representing using Radiance only (RGB), the middle shows a high resolution RGD ( horizontal 0.2 degree and vertical 0.65 degree ), and the bottom shows a low resolution RGD (horizontal 0.8 deg/sample and vertical 1.31 deg/sample). The images represent the same scene, but the B color channel is replaced with data from the depth map.  When depth is small, the images appears yellow (object is near) and when depth is large blue is large (object is far). Consequently, this image appears to be a gradation of more yellow to less yellow as a function of distance. The RGD images are a convenient representation to use as the input to the ResNet, which is designed for a radiance camera (RGB).

\Figure[t!](topskip=0pt, botskip=0pt, midskip=0pt)[width=3 in]{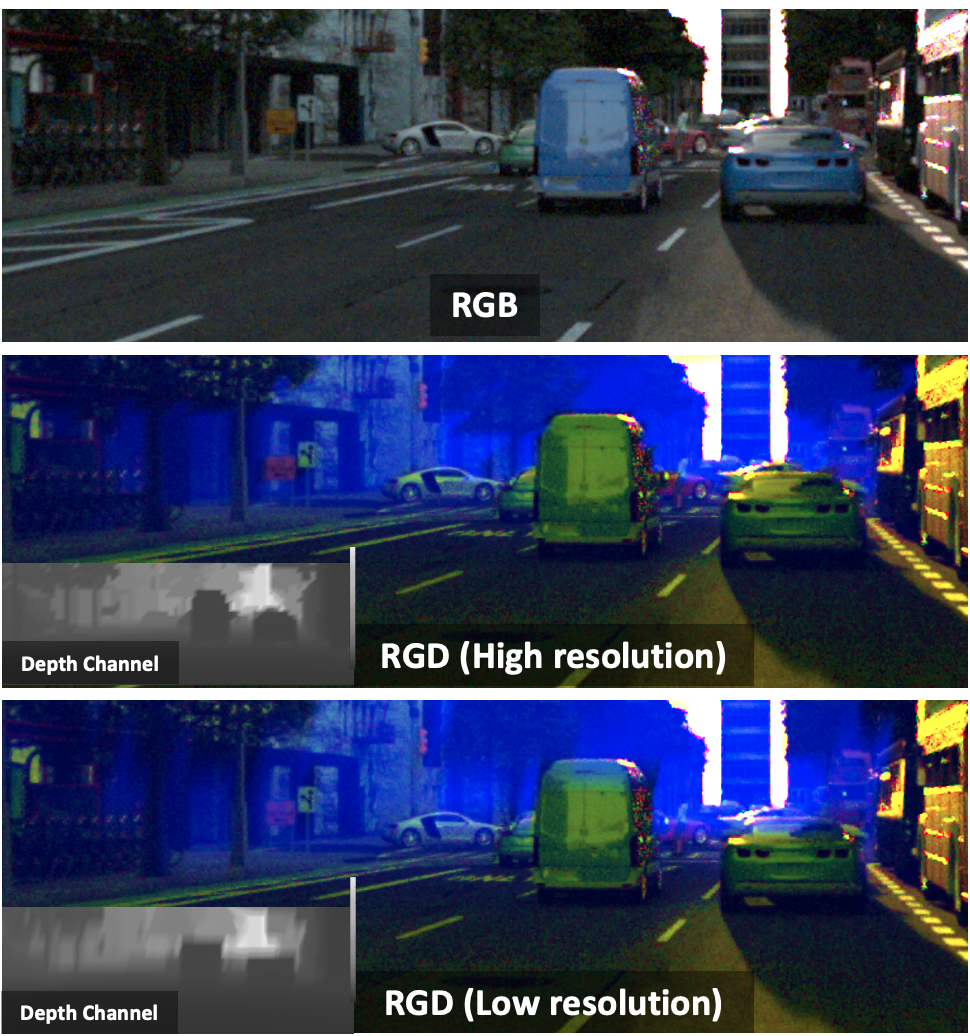}
{\ Simulated scenes illustrating the radiance image (RGB, top row) or a combined radiance and depth map (RGD, middle and bottom row). In the RGD representation the intensity of the nominal blue channel codes depth.  The depth map, sampled at a lower resolution than the image, is shown by the monochrome insets at the lower left of each image. For display we linearly interpolate the low resolution depth data to the RGB resolution. Because the visual system has low spatial resolution to short-wavelength light, the RGD images appear similar to one another despite the difference in depth sampling density. 
\label{fig6}}

\subsubsection{Evaluating AP with RGD inputs}
The data in Figure \ref{fig7} show the effect on average precision when the ResNet was trained with simulated data either for a conventional radiance camera (RGB), depth data (D), or the combination of radiance and depth data (RGD).  The average precision was evaluated at two different depth resolutions. In both cases combining radiance and depth outperforms radiance or depth alone.  The improvement is particularly significant when combining low-resolution depth data with the radiance image.  In that case the depth alone AP is about 75\%, the RGB alone is about 81\%, and the combination is about 86\%.

\Figure[t!](topskip=0pt, botskip=0pt, midskip=0pt)[width=3 in]{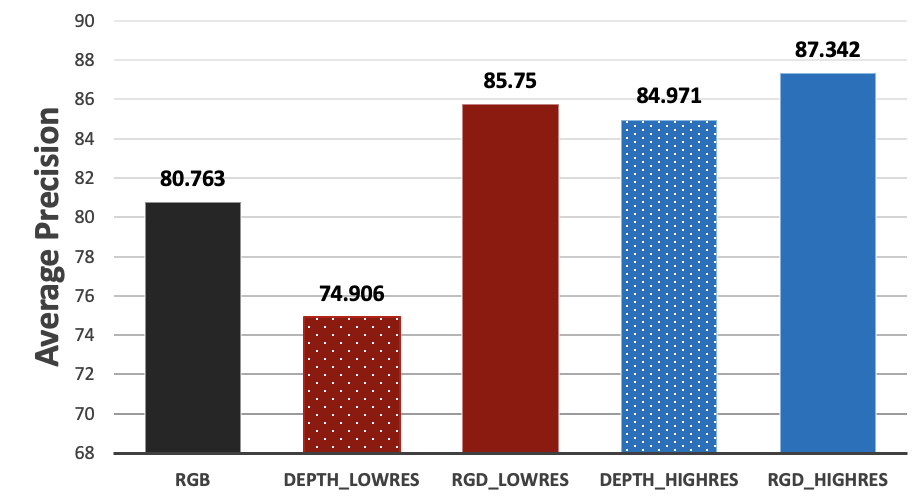}
{\ Average precision for vehicle detection using radiance alone (RGB), depth alone, or a combination of radiance and depth (RGD). The simulations were performed using low (red) or high (blue) depth spatial sampling resolution. The RGB spatial sampling  (black) was 1920 x 650. When  RGB resolution is reduced to that of the high resolution depth data, AP falls to 67\%. 
\label{fig7}}

\Figure[t!](topskip=0pt, botskip=0pt, midskip=0pt)[width=3 in]{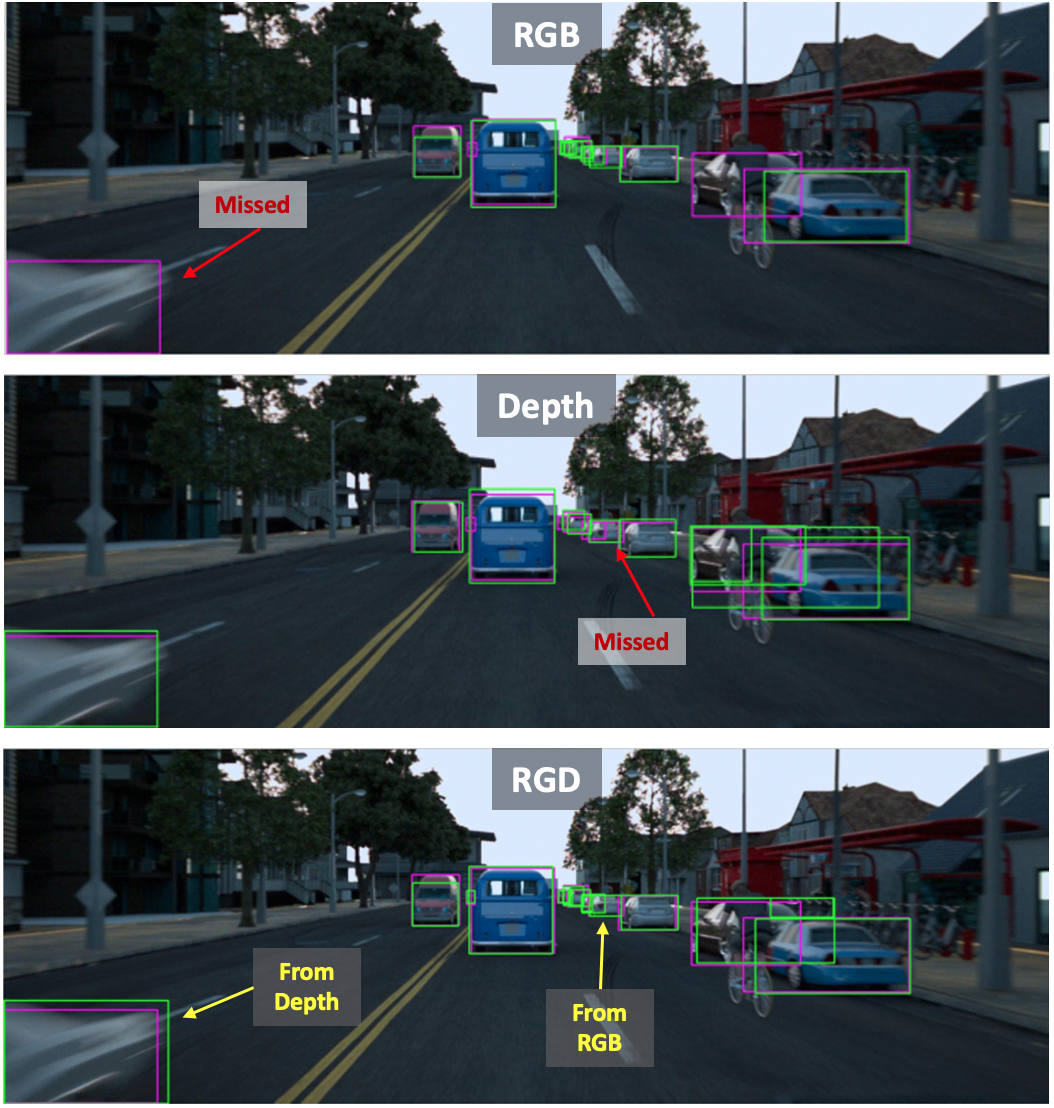}
{\ Vehicle detection examples on a ResNet trained with radiance, depth or both. The three images show the same scene with ResNet labeled vehicles (green) and ground-truth (purple). The top image shows the network trained on radiance alone (RGB), the middle trained on depth alone (horizontal 0.2 deg/sample, vertical 0.33 deg/sample), and the bottom trained on combined radiance and depth data (RGD).  Vehicles that are missed in the RGB trained (purple, top) network differ from the vehicles missed in the depth map (puple middle). The ResNet, trained on the combination of radiance and depth succeeds in both cases. 
\label{fig8}}

Figure \ref{fig8} shows specific examples in which the RGD sensor achieves better results than either radiance or depth alone.  Notice that the image includes a nearby vehicle that is moving, and thus its radiance image is blurred.  The vehicle is missed in the radiance image (RGB), but it is identified correctly in the depth map.  The distant vehicles at small resolution span very few sample points in the depth map and are missed in the depth image, but the high resolution RGB data detect the distant vehicles well. The vehicles are correctly identified in both cases when using the combined (RGD) data.

\subsection{Real-world data}
The simulation results are clear: combining radiance and depth information outperforms radiance or depth alone.  In this section, we ask whether we find the same pattern of results using publicly available radiance and depth data provided by Waymo \cite{Sun2019-bl}. Figure \ref{fig9} shows renderings of the combined radiance and depth data (RGD) that we constructed from that dataset.

\Figure[t!](topskip=0pt, botskip=0pt, midskip=0pt)[width=3.3 in]{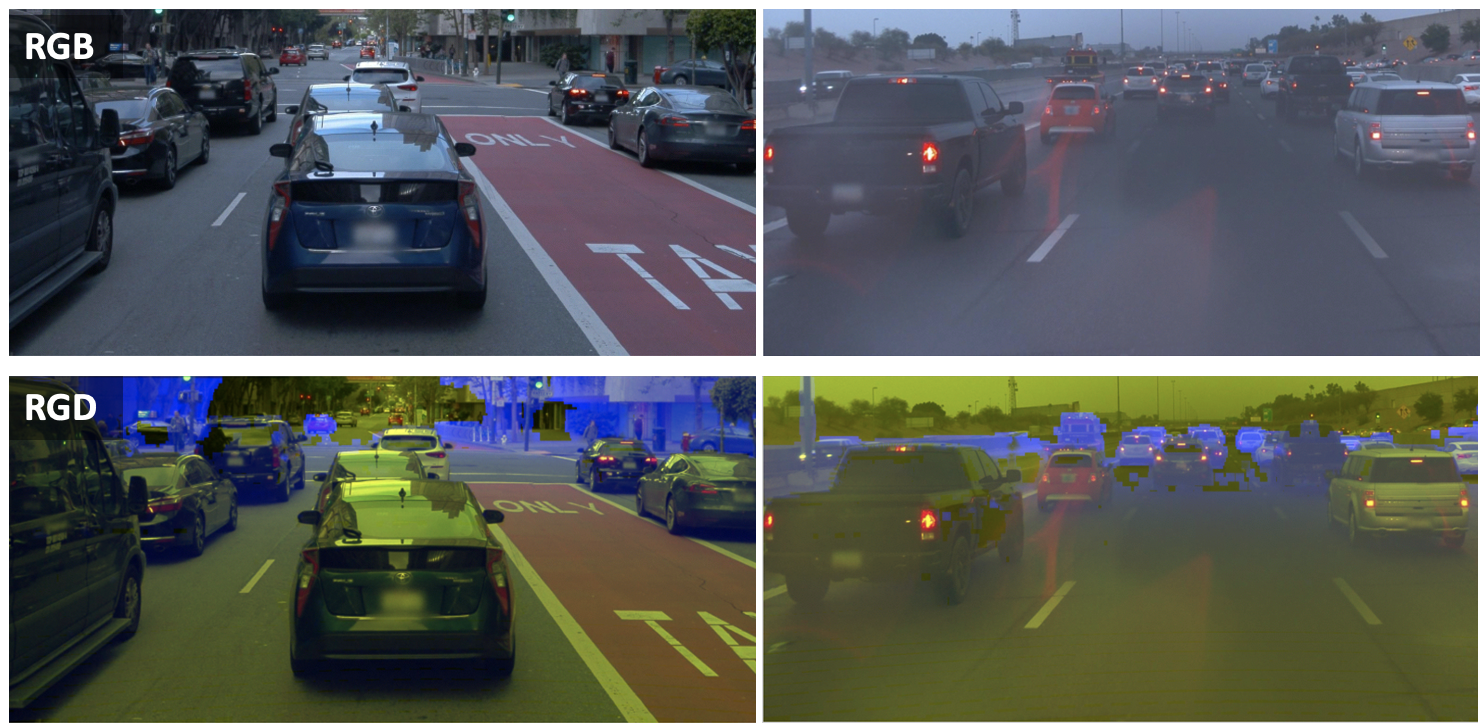}
{\ The two images at the top are radiance images from the Waymo dataset. The data set includes corresponding depth data. We combined the radiance and depth information into the two RGD images rendered at the bottom.
\label{fig9}}

Analyses using the Waymo dataset confirm the simulation findings. Training the ResNet on the RGB radiance data, on a monochrome channel alone, or the depth map alone, resulted in an average precision of around 76\% (Figure \ref{fig10}).  Combining the radiance and depth information (RGD)  increased the average precision to 81\%,  higher than either radiance or depth alone.

Similarly, as we observed in simulation, the depth map performs well even when its spatial density  is relatively low. In the Waymo dataset the spatial samples of the depth map comprise only 1\% of the spatial samples of the radiance (RGB) data.  Yet, the average precision levels using the two types of data are almost equal.  

The examples in Figure \ref{fig11} confirm that the increase in performance based pn the RGD input arises because the combined ResNet can be trained to detect information in both types of data, radiance and depth, taking advantage of the complementary strengths of the two types of measurements.  

\Figure[t!](topskip=0pt, botskip=0pt, midskip=0pt)[width=3 in]{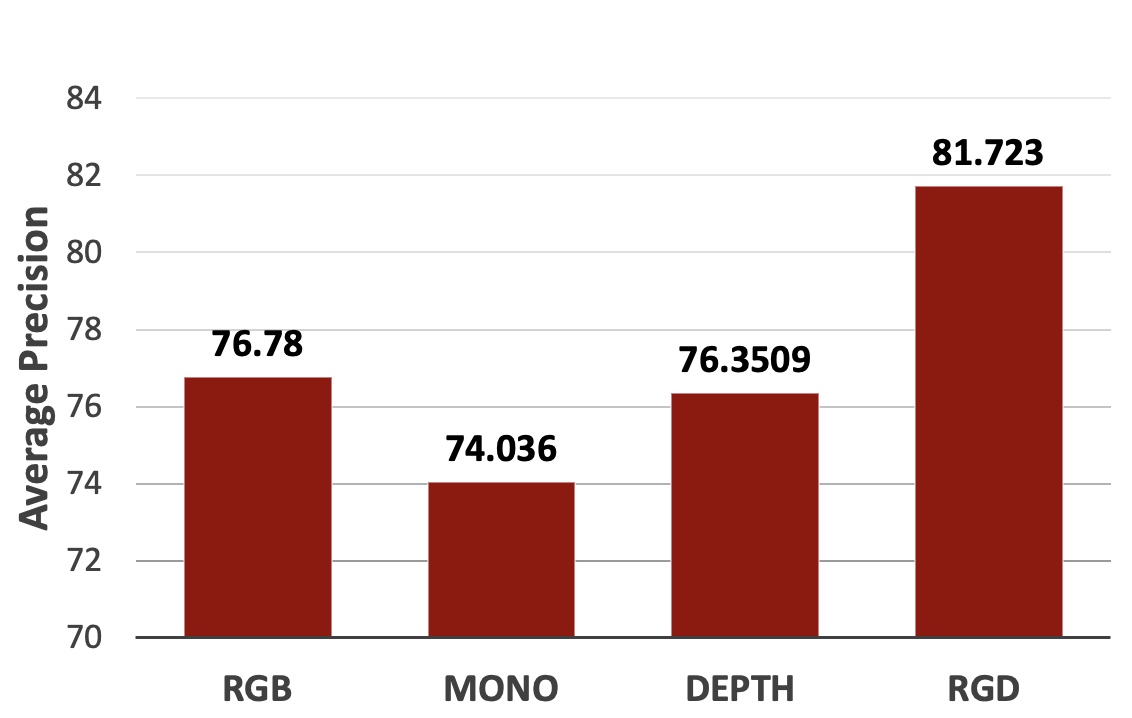}
{\ Calculations using the Waymo dataset confirm the simulations:  AP on the combined radiance and depth data is higher than either alone.  Training with the RGB data (1920x743) reached an AP of nearly 77\%. Simplifying the radiance data to a single luminance channel (MONO) decreases the AP by a little more than 2\%. The AP of a network trained on depth alone is similar to the AP of the RGB data, even though the depth data contain many fewer spatial samples. The ResNet trained on the combined radiance and depth information (RGD) exceeds that of the RGB and the depth by about 5\%.
\label{fig10}}

\Figure[t!](topskip=0pt, botskip=0pt, midskip=0pt)[width=3 in]{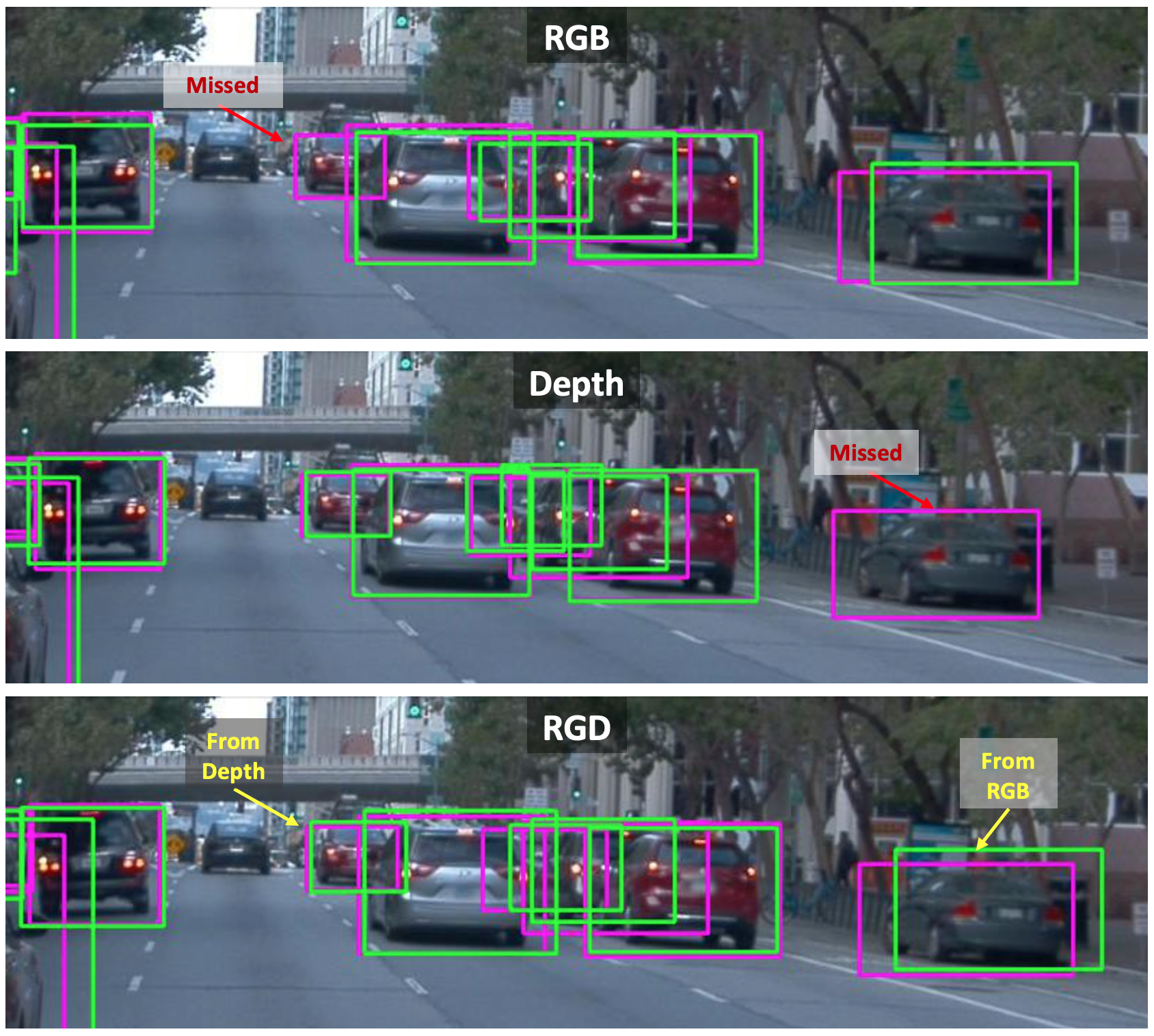}
{\ Vehicle detection examples from the Waymo dataset on a ResNet trained with radiance, depth or both.match the simulation analyses. Vehicles that were missed in RGB but found in depth, are also found in RGD.  Similarly, vehicles that were found in RGB but missed in depth are found in RGD. Box outline colors as in Figure \ref{fig8}. 
\label{fig11}}

\section{Discussion}
\subsubsection{Sensor measurement technologies}
The simulations quantify the increase in average precision  of a system that combines radiance and depth information compared to a system with either alone. Because the advantages can be significant it is worth considering the practical challenges of developing an integrated sensor that accurately measures co-registered radiance and depth images under driving conditions. 

\textit{Structured light.} Current technologies that simultaneously acquire radiance and depth information often use structured light (e.g., Kinect, Real Sense from Intel; Kinect style RGB-D type cameras).  \cite{Geng2011-og}. These  systems illuminate a scene with a known spatial pattern and measure the returned image, inferring depth from the difference between the known illumination pattern and the measured image.  Such technology is effective in certain contexts, but it is not appropriate for typical driving conditions.  

\textit{Time-of-flight.} LiDAR systems typically sweep a laser light through the scene and measure the time-of-flight using a small array of avalanche photodetectors. Single-photon avalanche detectors (SPADs) can be built in CMOS are viable as time-of-flight sensor arrays. Canesta also developed a time-of-flight system based on a special purpose CMOS image sensor. Quantum image sensor technology, also called a ‘jot’, uses a technology that is between the CMOS SPAD and conventional radiance detector. Gated imaging sensor\cite{Gruber2019-ax} systems also determine depth based on time-of-flight. These systems emit a very brief near-infrared illumination pulse and precisely control (gate) a sequence of very brief electronic exposure times. The duration of the pulse, the distance to the object, and timing of 3-5 exposure pre-determined exposure times produces a pattern of photon absorptions. The relative response from the several exposures can be used to estimate the distance to a scene object.

The circuit requirements for avalanche detectors and specialized time-of-flight circuits differ significantly from the circuits used to make radiance measurements. It might be possible to integrate both circuits on a single dye. With appropriately configured optics, such a chip could provide spatially aligned radiance and depth maps. Prior to implementing such a new detector, it would be possible to evaluate different designs using image systems simulation.  

\textit{Stereo pairs and camera arrays.} Depth can also be estimated using a stereo pair or array of cameras. This approach relies only on radiance and  would not require integrating radiance and time-of-flight technology. But the simulations demonstrate that a key limit of the radiance data include dynamic range and blur, and these problems are not solved by a system based only on radiance cameras. Furthermore, the simulations show that the spatial pattern of the missing depth information matters; when estimating depth from stereo pairs or arrays the spatial pattern of the missing depth information will be highly structured and very different from the missing information using LiDAR systems.

It will be useful to assess whether the same high level of vehicle detection can be obtained using depth derived from stereo.  The quality of stereo depth estimation algorithms continues to improve, and it may be that this can be an effective approach \cite{You2019-ex}.  Similarly, the quality of depth information obtained from SPAD arrays has generally been inferior to the data from standard LiDAR using avalanche detectors. There is promising research that seeks to improve the SPAD depth estimation by combining the time-of-flight data with radiance data (\cite{Sun2020-wg}). The simulations in this paper suggest that low resolution SPAD inputs that are properly aligned to the radiance data may be a useful approach to finding vehicles.

\subsection{Depth and radiance representations}
Investigators have used a variety of approaches to represent radiance and depth information. For example, some investigators convert RGB-Depth information into the format of height/horizontal disparity and angle (HHA \cite{Gupta2014-ux}). Others keep the depth and radiance information separate through multiple input stages, allowing them to converge only many layers deep in the network \cite{Chen2018-dj}.  Keeping the two sources of data distinct also permits the system to use very different formats for representing the radiance and depth information.

\textit{Point clouds and depth maps.} The depth maps and 3D point cloud formats are equivalent in the sense that there are transformations that convert precisely between them. Yet, for some applications point cloud representations may be advantageous. For example, \cite{Wang2018-vu} report that point clouds perform better for determining the 3D bounding box of a vehicle. The task we analyze is identifying 2D bounding boxes, and for this goal the depth map format improves performance significantly. Future experiments should explore how the effectiveness of the RGD representation for identifying 3D bounding boxes.
 
\subsection{Neural network architectures for data fusion}
In prior work, authors developed deep learning architectures to combine radiance and depth information \cite{Sun2020-bf,Chen2018-dj,Hazirbas2016-is,Ophoff2019-wo}.  The literature includes numerous innovative approaches for combining depth maps or point clouds with radiance data (Figure \ref{fig12}A-D).  Many of these architectures initiate the network by keeping the two modalities in separate channels and combining information from the distinct channels at layers that are deep within the network.  One paper systematically examined which network layer would be optimal for combining the two data streams (Fig 13E) \cite{Feng2019-sm} and concluded that performance is quite similar if one chooses early, middle, or late fusion architectures.

\Figure[t!](topskip=0pt, botskip=0pt, midskip=0pt)[width=3.35 in]{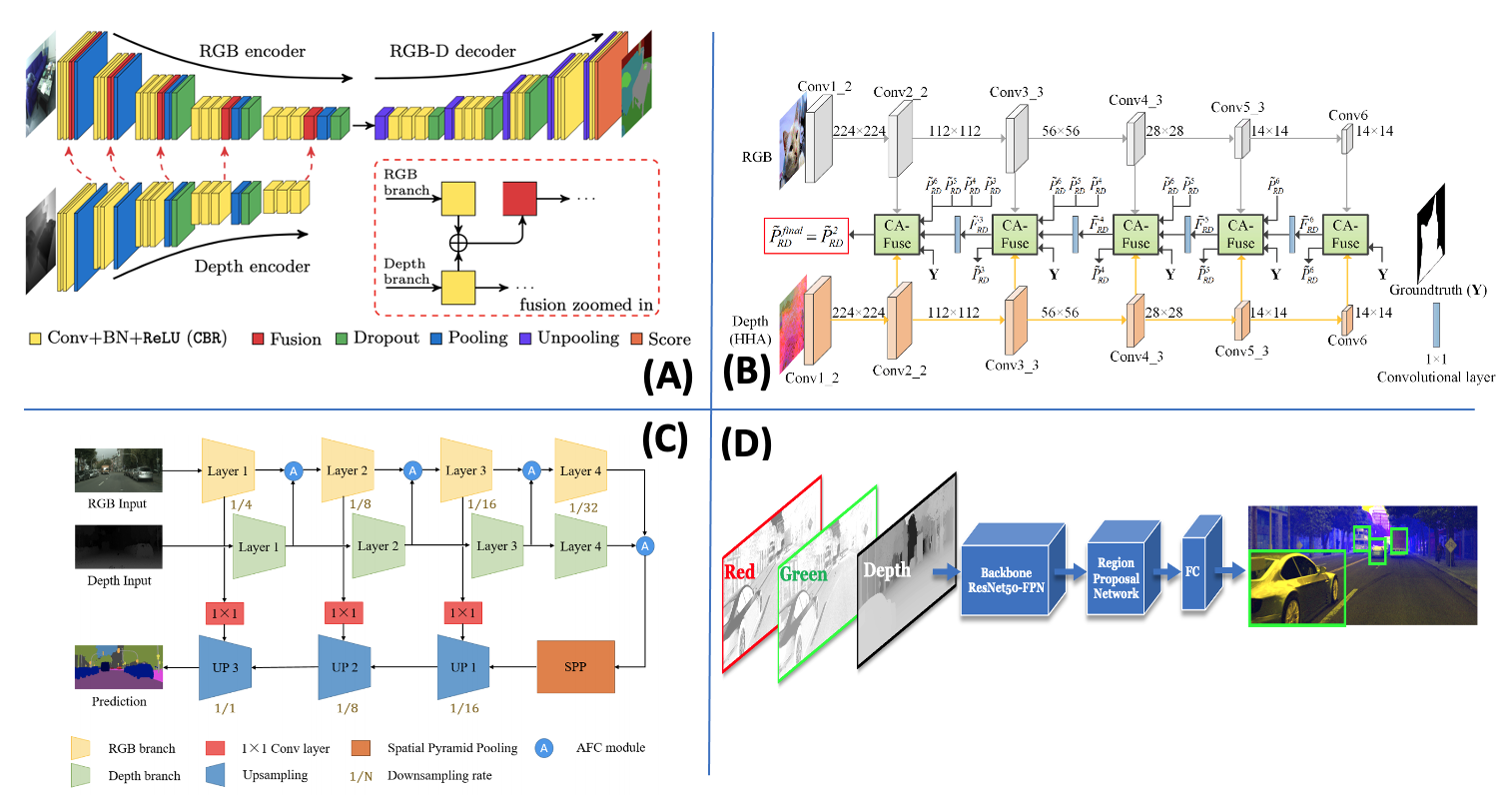}
{\ Sensor fusion architectures.  Panels (A-C) are overviews of the network architectures from three papers that fused radiance and depth images. Panel (D) is the ResNet architecture used in this paper that uses RGD as the fused input. (A) (Copied from Hazirbas et al., 2016)\cite{Hazirbas2016-is} (B) (Copied from Chen et al.2018)\cite{Chen2018-dj} (C)(Copied from Sun et al.2020) \cite{Sun2020-bf} (D) Ours: RGD.
\label{fig12}}

The simulations we describe here use modern CNN methods \cite{He2016-au} that take aligned depth and radiance images as input. The depth maps are combined with the radiance data at the earliest stage: the CNN input channel. Combining the data at the input, it is very straightforward to add a region proposal network \cite{Gkioxari2019-mw}. This architecture has been highly optimized; in some cases such a network performs at levels that reach optimal \cite{Reith2019-wy}. At this time, we see no reason to use a more complex architecture.

\section{Conclusion}
The analyses of simulations and empirical data quantify the value of depth information for vehicle detection. The results shows that after equating for spatial resolution, depth information is at least as valuable as radiance information.  When depth information is only available at the low spatial resolution, combining depth and radiance by inserting the depth map into an image input channel increases the average precision of vehicle detection substantially. We demonstrated this improvement using ISETAuto simulation, and we confirmed the finding using empirical data.  

The advantage of combining radiance and depth information can be explained by the fact that the two modalities have complementary weaknesses.  The depth information is acquired with extremely short duration exposures that limit the impacts of blur in moving targets. Also, depth information is less vulnerable to the dynamic range limits of cameras. Conversely, cameras have a spatial resolution advantage over LiDAR devices, and they are necessary for measuring some important information such as road markings, signs, and traffic signals. These observations suggest that there may be performance advantages for an integrated sensor that provides aligned radiance and depth images as an input to a CNN for vehicle detection.

{\small
\bibliographystyle{ieee}

\bibliography{access}
}
\titlepgskip=-15pt
\begin{IEEEbiography}[{\includegraphics[width=1in,height=1.25in,clip,keepaspectratio]{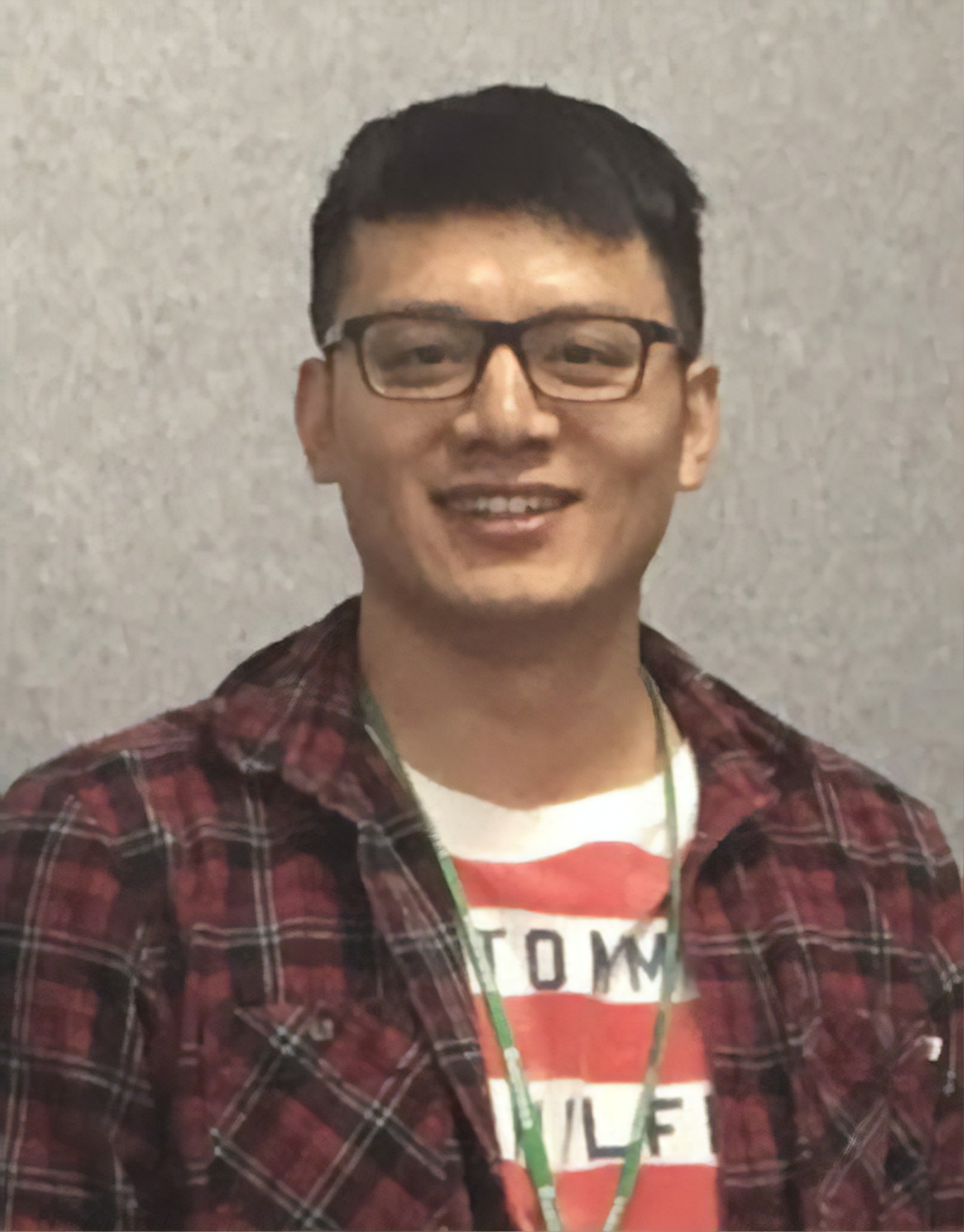}}]{Zhenyi Liu} received his MS in Electrical Engineering at Ulsan National Institute of Science and Technology, UNIST (2015), Korea. He is currently a PhD candidate in Automotive Engineering at Jilin University, China  (2016-present). Zhenyi was a Visiting Student Researcher at Stanford University (2017-2019). His research interests focus on machine perception systems for autonomous vehicles such as cameras and lidar.
\end{IEEEbiography}
% \vskip 0pt plus -2fil
\vskip -2\baselineskip plus -1fil
\begin{IEEEbiography}[{\includegraphics[width=1in,height=1.25in,clip,keepaspectratio]{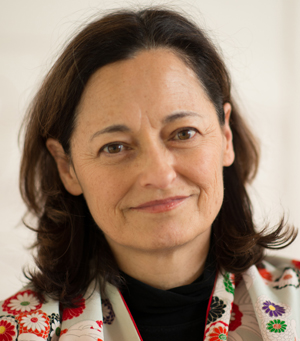}}]{Joyce Farrell} is a Senior Research Engineer and Lecturer in the Department of Electrical Engineering; she is the Executive Director of the Stanford Center for Image Systems Engineering. Dr. Farrell co-founded ImageVal Consulting and has more than 20 years of research and professional experience  working at a variety of companies and institutions, including the NASA Ames Research Center, New York University, the Xerox Palo Alto Research Center, Hewlett Packard Laboratories and Shutterfly.
\end{IEEEbiography}
\vskip -2\baselineskip plus -1fil
\begin{IEEEbiography}[{\includegraphics[width=1in,height=1.25in,clip,keepaspectratio]{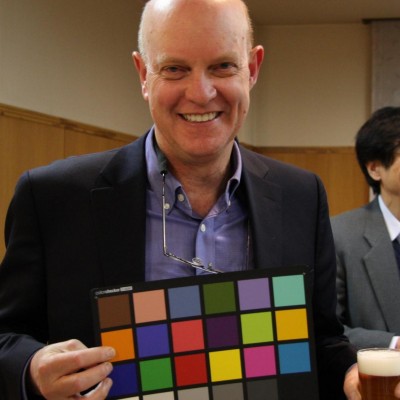}}]{Brian A. Wandell} is the first Isaac and Madeline Stein Family Professor. He joined the Stanford Psychology faculty in 1979 and is a member, by courtesy, of Electrical Engineering, Ophthalmology, and the Graduate School of Education. He is Director of Stanford’s Center for Cognitive and Neurobiological Imaging and Deputy Director of Stanford’s Neurosciences Institute. Wandell’s research centers on vision science, spanning topics from visual disorders, reading development in children, to digital imaging devices and algorithms for both magnetic resonance imaging and digital imaging.
\end{IEEEbiography}

\EOD
\end{document}